%% file: template.tex
\documentclass{article}
\pdfoutput=1

\usepackage{arxiv}
\usepackage{tabularray}
\usepackage[utf8]{inputenc} % allow utf-8 input
\usepackage[T1]{fontenc} % use 8-bit T1 fonts
\usepackage{hyperref} % hyperlinks
\usepackage{url} % simple URL typesetting
\usepackage{booktabs} % professional-quality tables
\usepackage{amsfonts} % blackboard math symbols
\usepackage{nicefrac} % compact symbols for 1/2, etc.
\usepackage{microtype} % microtypography
\usepackage{graphicx}
\usepackage{doi}
\usepackage{array}
\usepackage{chngcntr}
\usepackage{booktabs}
\counterwithin{figure}{section}
\counterwithin{table}{section}

\title{Pressmatch: Automated journalist recommendation for media coverage with \textit{Nearest Neighbor} search}

\author{
\href{https://orcid.org/0000-0001-9337-8552}
{\includegraphics[scale=0.075]{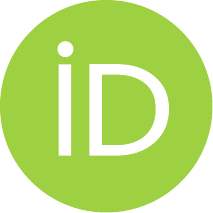}
\hspace{1mm}Soumya Parekh}\\
K. J. Somaiya College of Engineering\\
Mumbai, India\\
\texttt{soumya.parekh@somaiya.edu}\\
\And
\And
\And
Jay Patel\\
Specrom Analytics\\
Ahmedabad, India\\
\texttt{patel.jay@specrom.com}\\
}

\date{}

\hypersetup{
pdftitle={Pressmatch: Automated journalist recommendation for media coverage with Nearest Neighbor search},
pdfsubject={cs.LG},
pdfauthor={Soumya Parekh, Jay Patel},
pdfkeywords={Nearest Neighbors, Multilabel Classification, TF - IDF Vectorizer, Text Analytics, Data Mining},
}
\begin{document}
\maketitle
\begin{abstract}
Slating a product for release often involves pitching journalists to run stories on your press release. Good media coverage often ensures greater product reach and drives audience engagement for those products. Hence, ensuring that those releases are pitched to the right journalists with relevant interests is crucial, since they receive several pitches daily. Keeping up with journalist beats and curating a media contacts list is often a huge and time-consuming task. This study proposes a model to automate and expedite the process by recommending suitable journalists to run media coverage on the press releases provided by the user.
\end{abstract}
% keywords can be removed
\section{Introduction}\label{Introduction}
Press releases today are invaluable tools to get the word out about new products and services, which drive engagement, traffic, and audience reach. Media pitches are often used to persuade journalists to run coverage on those news stories, effectively increasing their popularity. Finding the right journalists to run those stories makes a real difference in garnering interest in the product's offerings.
\newline
\newline
This can often prove to be tedious, as journalists are frequently flooded with several such pitches. Additionally, different journalists often work on different beats or news topics, which makes it essential to find the right contacts. Hence, pitching press releases to the right journalists with relevant interests makes them more likely to run media coverage on press releases. This can often be a dealbreaker when it comes to the popularity and revenue of the pitched products.
\newline
\newline
However, finding the most suitable journalists for this process often becomes an uphill task without the right contacts and a media list. Moreover, staying updated with the recent interests of journalists and the trends they cover can often be a tedious task. Our work aims to automate this process by leveraging text analytics.
\newline
\newline
Text mining, or text analytics, is a branch of artificial intelligence that uses natural language processing techniques to extract useful insights from large amounts of text data. This process involves several steps that transform real-world, unstructured data into a structured, normalized representation that can be fed into predictive models or analyzed to uncover hidden patterns that can provide actionable insights to decision-makers. Text mining has a variety of applications ranging from fraud detection, personalized advertising, risk management, and so on.
\newline
\newline
This study proposes the use of text mining techniques combined with predictive models to automate and speed up the process of finding relevant journalists who might be interested in covering potential press releases by taking into account their interests. It also compiles media contact information that enables users to quickly find and get in touch with the most suitable journalists to take point and optimize engagement on their press releases. Figure 1.1 outlines and summarizes the approach followed in this paper.
\begin{figure}[h]
\centering
\includegraphics[width=1\linewidth]{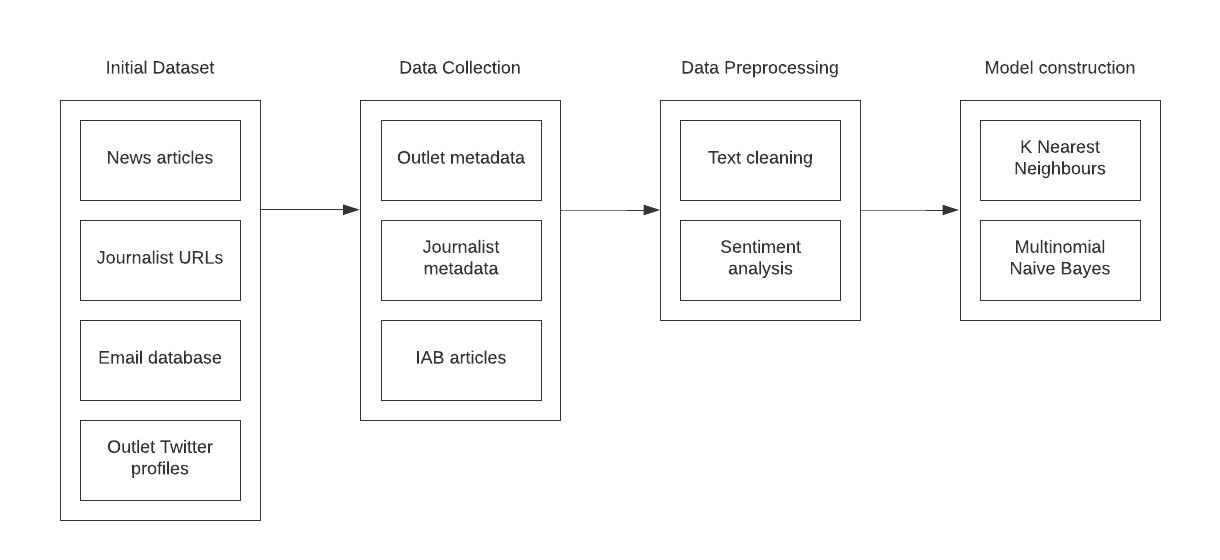}
\caption{Workflow to automate journalist recommendations for media coverage.}
\label{fig:Figure 1.1}
\end{figure}
\section{Related Work}\label{Related Work}
In the previous decade, a lot of advances have been made in leveraging machine learning capabilities with regard to news articles and press releases. Significant research has been focused on applying text processing techniques on publicly available news datasets to give rise to methods for the detection of fake news, entity modeling, news summarization as well as analyzing competitor strategy from press releases.
\subsection{Fake News Detection}\label{Fake News Detection}
Due to the ease of distribution and abundance of information today, it has become an increasingly harder task to filter out credible information from rumors, hoaxes, and questionable news. Fake news detection aims to achieve this task using machine learning algorithms including linguistic and non-linguistic \cite{1, 2, 3} clues to interpret the frequency and usage of certain words and phrases. These approaches can be useful in segregating the most relevant and genuine news content from the rest.
\subsection{Topic Modeling}\label{Topic Modeling}
Most news articles express their ideas by including a variety of references to different entities or topics like people, places, and objects. Topic modeling \cite{4} discovers topics or entities within different news articles by methods like Latent Dirichlet Allocation \cite{5}, Correlated Topic Model, and Probabilistic Latent Semantic Indexing. The modeling is usually probabilistic, by modeling topics over the existing vocabulary in the form of probabilistic distribution.
\subsection{News Summarization}\label{News Summarization}
News summarization summarizes articles that are otherwise quite long, to assist readers in capturing the gist of the content more quickly. Extractive or abstractive \cite{6, 7} techniques can be employed to paraphrase long documents by generating summaries of various sections in the text. Recent research also employs attention-based models to improve results.
\section{Proposed Work and Methodology}\label{Proposed Work and Methodology}
\subsection{Dataset Description}\label{Dataset Description}
Most of the datasets used in this paper come from public sources like Muckrack and Twitter, which are scraped to collect information in a methodical and organized manner using Python libraries like BeautifulSoup and Requests.
\subsubsection{Article Collection}\label{Article Collection}
A collection \cite{8} of almost 267k news articles authored by various journalists containing the article topic, title, description, website links, and outlets they mainly write for is used as a starting point to get an insight into the relevant interests of those journalists. Figure 3.1 depicts the distribution of articles into the 5 main topics - sports, politics, entertainment, tech, and business.
\begin{figure}[h]
\centering
\includegraphics[width=0.5\linewidth]{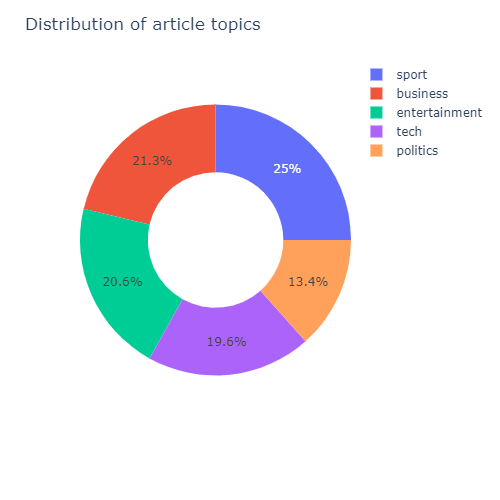}
\caption{Article distribution across topics.}
\label{fig:Figure 3.1}
\end{figure}
\subsubsection{Muckrack Weblinks}\label{Muckrack Weblinks}
Muckrack is an online platform that has dedicated pages to various journalists along with a comprehensive summary of their profiles. Web scraping was used to save all the links from the sitemap, thus adding additional means of looking up the journalists and contacting them.
\subsubsection{Email Database}\label{Email Database}
A database of over nearly 2500k various emails along with first and last names is utilized to find potential points of contact for journalists.
\subsubsection{Twitter Profiles}\label{Twitter Profiles}
Publicly available Twitter information about media outlets is scraped to gather data on their follower counts, which is used to verify the credibility of various outlets.
\subsection{Data Collection}\label{Data Collection}
The initial article dataset contains authorship information on various news articles, with many journalists often collaborating on various pieces. Using this data, one can uncover article patterns for various journalists which can be leveraged to recommend individuals who might cover a press release based on relevant articles written in the past. Text processing libraries like NLTK \cite{9} are combined with powerful web scraping tools to provide end-users a means to contact these journalists - namely Muckrack profiles, Linkedin, and Twitter handles. Additionally, some information needs to be gathered for the media outlets as well, which includes statistics on their ranking, popularity, and Twitter followers. Finally, some news articles also need to be collected as an input to the classifier which can tag articles with various categories. Hence, the data collection process is divided into 4 major steps.
\subsubsection{Journalist Data}\label{Journalist Data}
For this paper, only journalists who have authored at least 10 articles across all media outlets are considered. The initial dataframe follows a few processing steps to separate journalists from combined authorship on articles and remove irrelevant keywords or tags from their titles to get a full name. Names formatted with bad Unicode are also removed. The most exhaustive information on author profiles is taken from Muckrack. Muckrack URLs are usually formed in the following format in most cases:
\newline
\newline
'https://muckrack.com/' + 'author\_firstname' + '-' + 'author\_lastname'
\newline
\newline
As a point of interest, the platform has an autocorrect service that redirects URLs with a few minor mistakes to the correct page on its own. One can try requesting web pages this way for individual journalists, but some URLs include some additional numbers or middle names. Hence, a sitemap is used in collaboration with the following string-matching techniques to link web pages correctly to the individual authors:
\begin{enumerate}
\item Levenshtein distance: This distance metric is computed by finding the minimum number of single-character edits to transform one word into the other. In this case, a string formed by the above method is compared with all strings in the Muckrack sitemap. A potential URL for any journalist has the least Levenshtein distance from the original string, which indicates a minimum number of changes required to transform one string into another \cite{10}. Match ratios are calculated on a scale from 1 to 100, where a higher value indicates the most probable Muckrack URL for a journalist.
\item Fuzzy matching with TF - IDF: TF - IDF or Term Frequency - Inverse Document Frequency is a fuzzy matching \cite{11} technique for large datasets. This technique counts the frequency of n-grams in the input strings and takes into account how commonly those strings occur in the dataset. This has a smoothing effect on the noise and gives less importance to those ngrams which occur more frequently. String matches are computed using similarity in string vectors, making it a faster and more efficient approach for this problem. Match confidence is calculated between 0 to 1, and for this paper, any matches with greater than 0.85 confidence are accepted as valid.
\newline
\newline
Fuzzy matching is favored over the previous approach for finding the most similar URLs \cite{12, 13} from the sitemap since the calculation of Levenshtein distance does not scale up well for larger datasets. Due to its quadratic time complexity, it takes a significantly greater amount of time as compared to TF - IDF matching.
\input{table_1}
\item Web search: For the rare cases in which neither of the two above approaches yields a valid match, we simulate a web search for the journalist's Muckrack profile Algorithmia's Google search service.
\end{enumerate}
Using the Muckrack URLs collected, Python's web scraping libraries are used to collect journalist beats. Beats are thematic topics that journalists usually cover in their work, making it easier to find individuals willing to run stories on specific fields of interest. Additionally, Linkedin, Twitter, and Facebook handles are also collected to establish a means of contact with them.
\subsubsection{Email Matching}\label{Email Matching}
To establish another way to get in touch with journalists, the email dataset is used to find matches with the list of authors. Since the number of email addresses is huge (2574k), finding matches for every individual can be a highly time-consuming process, even with both previous string-matching techniques. Hence, deduplication is performed using record linking, a technique used to link records from different data sources \cite{14}.
\newline
\newline
Here, the data sources used are the author names, the outlets they write for, and the list of email addresses. Among those multiple records, some of the information could be about the same entity. This would indicate that the records contain duplicate information. Deduplicating involves searching for and matching such duplicate records from multiple sources. Record linkage provides a way to reduce the number of computations involved by blocking columns on the first name. As we compare only records with the same first name, the time complexity decreases significantly.
\newline
\newline
Most emails are often of the form 'firstname' + 'lastname' + '@' + 'outlet.com', where the outlet is often the media outlet the journalist writes for the most frequently. Hence, record linkage is performed by creating strings in this format. Match thresholds vary from 0 to 1, with 0 being the least similar and 1 being the most similar. For this experiment, a threshold of 0.5 is selected. Matched emails are validated by an automated web search of the matched email address to see if the results link back to the journalist's Muckrack page.
\newline
\newline
This process results in 554 matches but yields only 79 absolutely correct matches based on the selected match threshold. True matches do have a higher match threshold than invalid and false matches, however, this experiment does not work too well in practice since the number of usable matches is not very significant.
\input{table_2}
\subsubsection{Outlet Data}\label{Outlet Data}
Amazon's Alexa Top Sites service allows developers to retrieve ranking and traffic data for a wide range of websites. The API is used to collect popularity statistics for the media outlets in our database, namely popularity, reach rank, and country rank. These statistics help evaluate the quality of work based on the rankings of the outlets journalists work for. Figure 3.2 depicts those metrics for the top 15 media outlets based on article count.
\begin{figure}[h]
\centering
\includegraphics[width=1\linewidth]{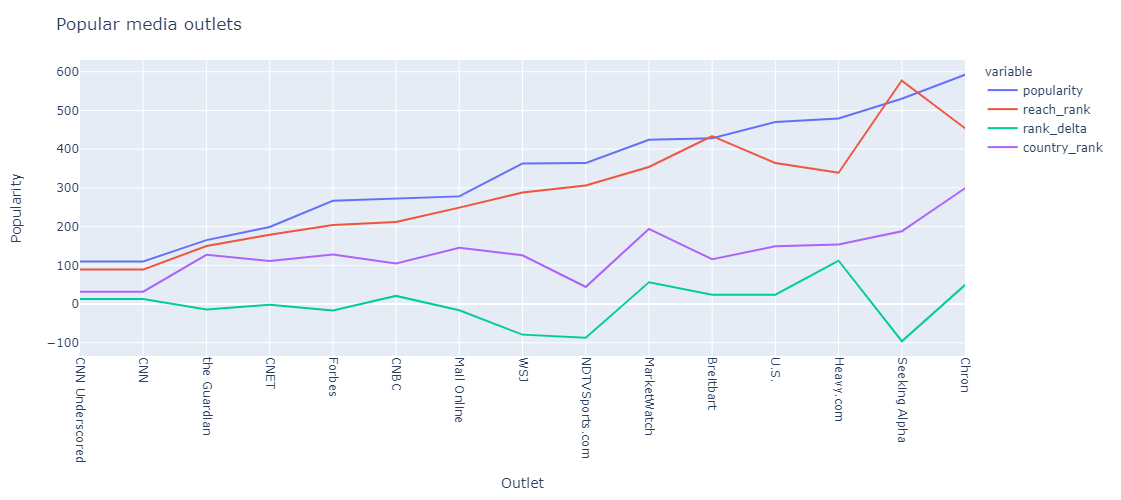}
\caption{Comparison of popularity metrics for the top 15 media outlets.}
\label{fig:Figure 3.2}
\end{figure}
\subsubsection{Articles for the IAB classifier}\label{Articles for the IAB classifier}
To create a model that classifies articles into different categories, IAB's content taxonomy system is used for text categories. 10 articles are scraped for each of the 1200 different class names using Python's web scraping libraries, which are then filled in for each of the 4 tiers, with Tier 1 being the least specific and Tier 4 being the most specific. Text is collected by 2 methods - by extracting all text content on the webpage using Python's Beautiful Soup, and by removing all boilerplate HTML from the page source.
\subsection{Data Preprocessing}\label{Data Preprocessing}
\subsubsection{Text Cleaning}\label{Text Cleaning}
Text data collected from the web is very messy in practice since raw strings contain punctuation, bad Unicode, and plenty of common words like articles, which do not contribute significantly to the results. Hence, it is hard to interpret the text collected unless it has been cleaned. To create models that pick up on the meaning of the words, all the text articles and the data collected for the IAB classifier are processed using the following steps by leveraging Python's libraries \cite{15}.
\newline
\newline
All bad Unicode is converted to clean, UTF-8 encoding \cite{16} after converting all text to the same case, i.e. lowercase. All the text is converted into words, or tokens, using a process called tokenization, and all punctuation signs are removed. We also remove stopwords or commonly occurring words like 'a', 'the', 'is', etc since those words do not contribute significantly to the meaning of the text. Each word is also converted into its root form, which is a technique called lemmatization. POS (part of speech) tagging is used on those words to better get the context of the words - such as noun, adjective, verb - which enhances the lemmatization process to transform tokens into their contextual root words.
\newline
\newline
Additionally, some of the text scraped from the internet also contains emojis. These characters are removed using regex patterns from the strings. For the model construction, all non-English text data is also filtered out.
\subsubsection{Sentiment Analysis}\label{Sentiment Analysis}
The Afinn lexicon \cite{17} is used to perform sentiment analysis \cite{18, 19} on all the journalist articles to get an insight into their writing profile. This process involves calculating the proportion of words in the articles that have positive and negative contexts. Using the Afinn lexicon for sentiment analysis \cite{20} yields the sum of valence scores for the article. Higher scores indicate a more positive connotation for the text, whereas lower scores indicate a more negative connotation.
\newline
\newline
Figure 3.3 depicts Afinn sentiments \cite{21} across all articles falling under the 5 main topics.
\begin{figure}[h]
\centering
\includegraphics[width=1\linewidth]{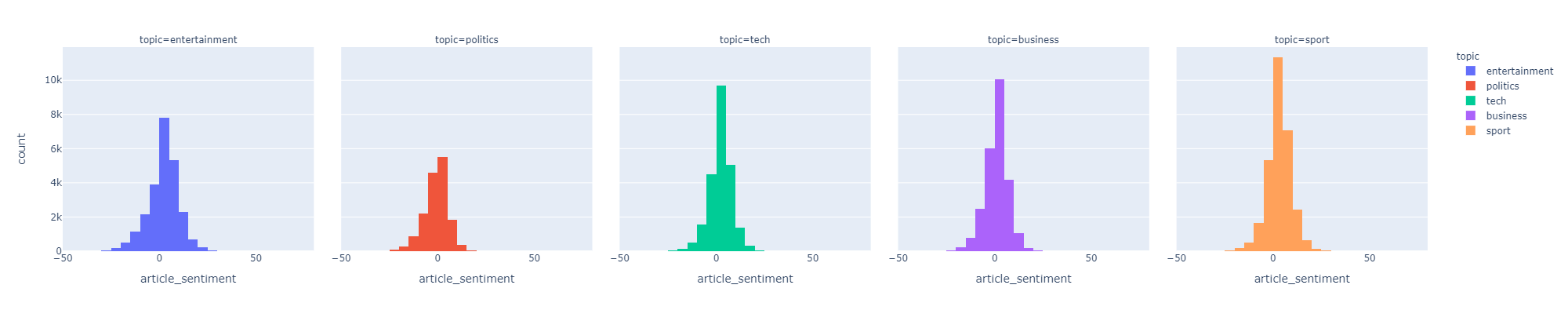}
\caption{Breakdown of valence scores using the Afinn lexicon across articles in the 5 main topics.}
\label{fig:Figure 3.3}
\end{figure}
The calculated sentiment scores and article counts can also be used to get the top journalists and mean sentiments per journalist for any media outlet as shown in figures 3.4 and 3.5.
\begin{figure}[h]
\centering
\includegraphics[width=1\linewidth]{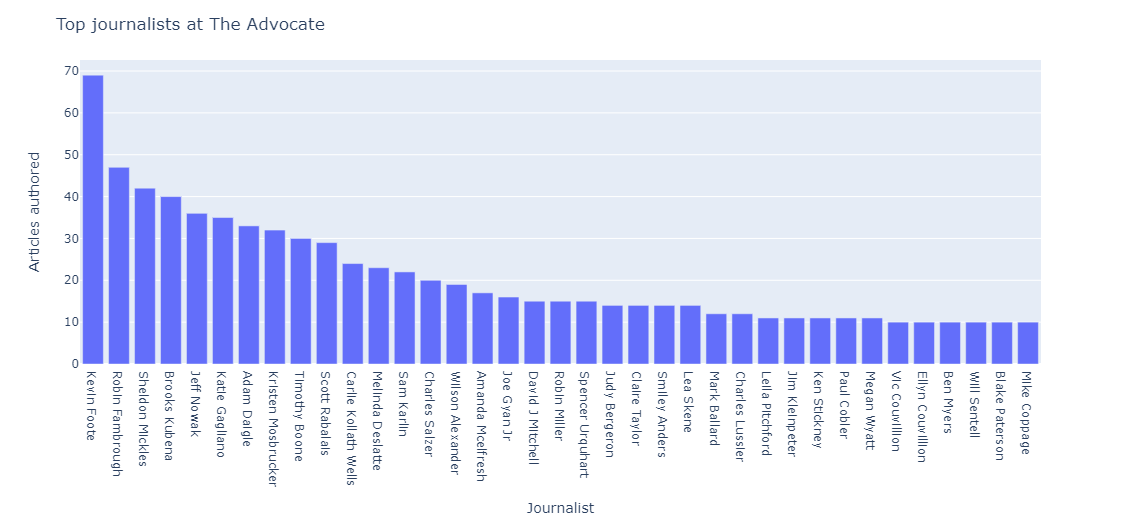}
\caption{Top journalists across media outlets ranked by number of articles authored.}
\label{fig:Figure 3.4}
\end{figure}
\begin{figure}[h]
\centering
\includegraphics[width=1\linewidth]{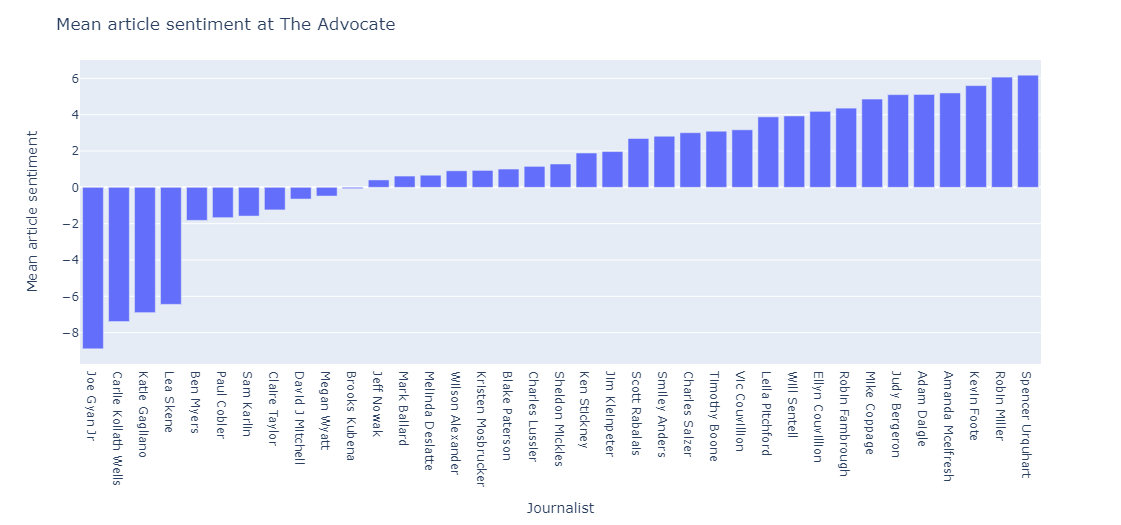}
\caption{Mean sentiments per journalist at \textit{The Advocate} computed using Afinn scores.}
\label{fig:Figure 3.5}
\end{figure}
\subsection{Model Construction}\label{Model Construction}
Our aim was to construct a model that can use a proposed press release as input which will return a list of potential candidates it can be pitched to. Journalists who have run stories on specific news beats in the past are more likely to run coverage of press releases on similar topics. Since the model takes into account recent writing activity, it recommends individuals who have worked on similar articles in the past, and thus have interests aligned to the press release in question. This helps make an informed decision about selecting journalists to pitch new press releases to.
\subsubsection{Nearest Neighbors Classifier}\label{Nearest Neighbors Classifier}
The proposed classifier accepts a text article as an input and returns the list of journalists who authored articles that are the closest, or most similar to it. To compute the closest neighbors to a given press release, there are a variety of distance metrics that can estimate the proximity between different data points. Points that are closer to each other with lesser distance have a higher similarity and vice versa \cite{22}.
\newline
\newline
To perform feature extraction from pieces of text, it must be transformed into a numerical representation to create word vectors. Count vectorizer and the TF - IDF vectorizer are 2 such methods to generate word vectors from a given corpus of text, ie. the news articles written by different journalists in our case. This paper uses the TF - IDF (or term frequency-inverse document frequency) for the given corpus \cite{23}.
\newline
\newline
TF stands for term frequency, which indicates the number of times a term t occurs in a document d. IDF, or inverse document frequency, represents the rarity of term t in all documents, which is calculated as log (all documents/number of documents in which the term t occurs). Words occurring frequently across documents do not have as much significance and have a lower IDF value, but rarer words have a higher value since the less frequently a term occurs, the more important it is \cite{24}. Thus, every word is assigned a weight W = TF * IDF. Terms occurring across all documents get assigned the weight of 0, whereas those occurring less frequently have higher weights.
\newline
\newline
The sparse TF - IDF matrix is constructed by finding the weights for all words in the corpus. There are various metrics to compute the distance between feature vectors, such as Euclidean distance, Manhattan distance, Hamming distance, and so on. This study uses cosine distance as a similarity metric between feature vectors. Table 3.6 depicts an example text article and Figure 3.6 graphs the TF - IDF weights for the 15 most significant words in the sample.
\input{table_3}
\begin{figure}[h]
\centering
\includegraphics[width=1\linewidth]{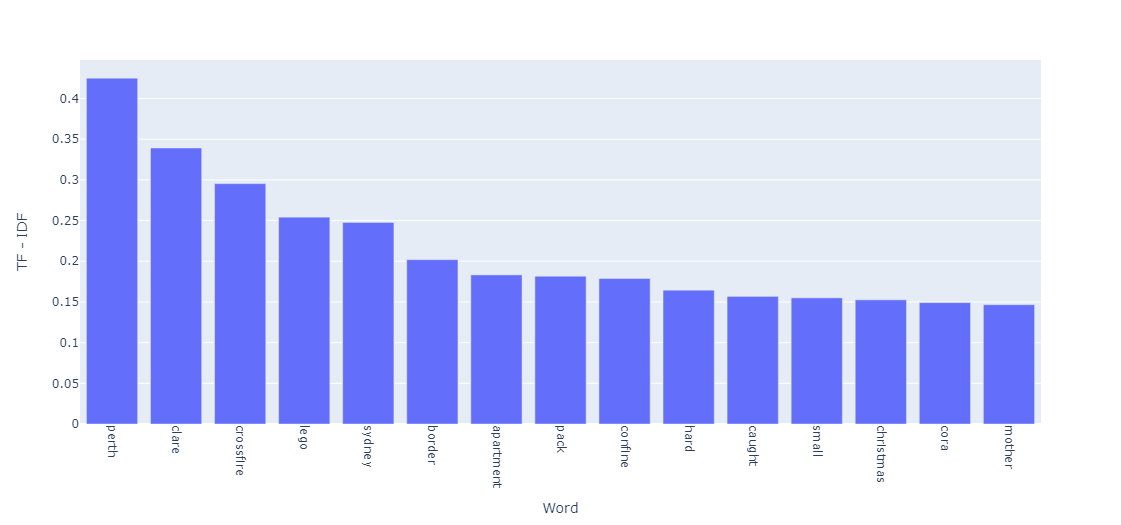}
\caption{Tf - IDF weights for the 15 most significant words from the article shown in Table 3.6}
\label{fig:Figure 3.6}
\end{figure}
When given an unlabeled press release as input, the model converts the text into a feature vector. The distance of this vector from all feature vectors in the set is computed. Similar documents have similar feature vectors and thus have a lesser distance between them. We then select the K Nearest (k = 5) vectors and look up the journalists who have authored the most similar articles when compared to the input text. Those names are suggested as potential candidates to run coverage on the given press release, along with their usual topics of interest, similar articles, and the outlets they appeared in. Any relevant contact information to get in touch with them such as Linkedin handles, email addresses, or Muckrack profile links are also provided.
\subsubsection{Custom IAB Classifier}\label{Custom IAB Classifier}
News articles can be further classified into successively refined categories or text tiers. The IAB Content Taxonomy tiers have 4 levels of depth, with each category further classified into subcategories. There are 38 main tiers, which include categories like Automotive, Books and Literature, Business and Finance, Careers, Education, and so on. For instance, a Tier 1 label for Education is further divided into Tier 2 labels such as Secondary Education, College Education. They are further divided into Tier 3 categories like College Planning and Standardized Testing, which contain a Tier 4 subcategory, Professional School. Figure 3.7 depicts this relationship.
\begin{figure}[h]
\centering
\includegraphics[width=1\linewidth]{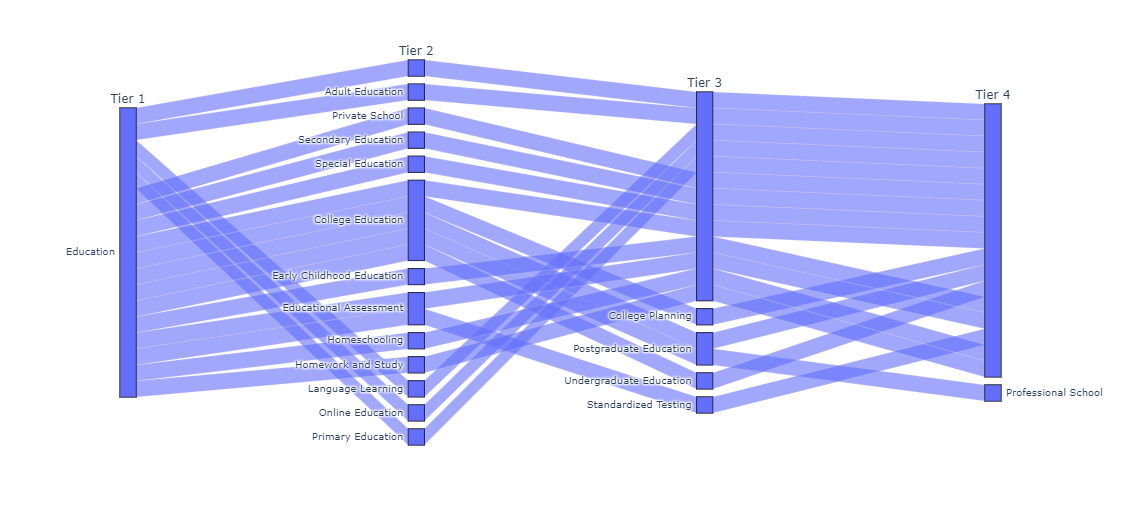}
\caption{Categories in the IAB Content Taxonomy tiers.}
\label{fig:Figure 3.7}
\end{figure}
Multilabel classification models are used to assign multiple labels to the given input. Thus, an IAB classifier is created to predict the 4 IAB text tiers for any given text. The classifier is trained on different columns of the obtained text and with the following different approaches:
\begin{enumerate}
\item One-Vs-Rest and Multinomial Naive Bayes:
The OvR or One-Vs-Rest technique splits one multilabel \cite{25} classification problem into multiple binary classification problems by creating multiple models. The models make predictions for every binary classification problem with certain probabilities, where the answer with the highest probability is used to make a prediction.
\newline
\newline
A multiclass classification problem is different from a multilabel classification problem since the former assigns a data point to one class out of multiple classes, whereas the latter can assign one data point to more than one label out of multiple classes. A Multinomial Naive Bayes \cite{26, 27} classifier uses a probabilistic algorithm that computes conditional probability with the following formula:
\newline
\begin{equation}
P(A | B) = P(A) * P(B | A) / P(B)
\end{equation}
\newline
P(B) refers to the prior probability of the text belonging to class B,
P(A) refers to the prior probability of the text belonging to class A,
P(B | A) refers to the probability that text belongs to class B given that it belongs to class A.
\newline
\newline
Hence, a Naive Bayes classifier is used along with the OvR classifier to solve the multilabel classification problem for predicting IAB content taxonomy tiers for any given text article.
\item K Neighbors Classifier:
A second alternative to performing multilabel classification is training a K Neighbors Classifier to predict IAB tiers for the text input. After converting text into feature vectors using the TF - IDF vectorizer, the model computes the top 3 closest neighbors to the piece of text given and assigns the labels to it based on the labels of the nearest neighbors.
\end{enumerate}
\section{Results and Conclusion}\label{Results and Conclusion}
The IAB classifier is trained separately using text obtained with both methods - text directly collected from the webpage (using Python's web scraping tools), as well as text collected by removing boilerplate HTML code from the webpage. Various metrics such as accuracy score, recall, weighted score, and precision are used to evaluate the performance of the classifier as shown in Table 4.1.
\input{table_4}
To test the model on new text inputs, we use information scraped from publicly available Twitter data of media outlets. Outlets with at least 3 journalists are considered. Only rows with several followers higher than the median and Alexa ranking lower than the median are used to get a good quality of outlets to test the performance on. All articles are requested for 1000 randomly selected journalists, out of which 5000 articles are randomly picked. All text is cleaned using Python's text processing libraries.
\newline
\newline
The IAB classifier is used to predict the Tier 1 labels for those articles, which are compared against the Muckrack beats of the journalist who authored the article. This is done using a mapping from beats to tiers. Only those news articles are selected which are authored by journalists having valid Muckrack beats, which do not include beats like Content Source Geo, Content Language, Content Source since these are generic beats that span a large variety of topics and hence cannot be assigned to only a few specific labels. 
\newline
\newline
For every news article, the author's beats are converted into a list of potential Tier 1 labels they could correspond to using the mapping. These labels are then compared to the actual prediction made by the IAB classifier, where the classification is marked as correct if there is at least one matching tier between the lists. An accuracy of 58.47 is obtained in this manner.
\bibliographystyle{unsrt}

\end{document}

%% file: table_1.tex
\begin{table}[ht]
\centering
\begin{tblr}{
cell{1}{1} = {r=5}{},
cell{1}{3} = {c=2}{halign = c},
hline{1, 6} = {-}{},
hline{2, 3} = {2-4}{},
}
{ 
Time taken\\(seconds)} & & Technique used &\\
& Matches & Levenshtein distance & Fuzzy Matching\\
& 1 & 71.21 & 4.95\\
& 10 & 811.23 & 54.54\\
& 100 & 9547.32 & 589.84\\\\
\end{tblr}
\caption{Comparison of times taken to compute matches using Levenshtein distance and Fuzzy Matching.}
\label{tab:Table 3.1}
\end{table}

%% file: table_2.tex
\begin{table}
\centering
\begin{tblr}{llll} 
\hline
& Invalid & False & True\\
\hline
Number of matches & 348 & 127 & 79\\
% \hline
Mean match threshold & 0.57 & 0.54 & 0.61\\
\hline\\
\end{tblr}
\caption{Match thresholds for true, false and invalid matches after performing record linking.}
\label{tab:Table 3.2}
\end{table}

%% file: table_3.tex
\begin{table}
\centering
\begin{tblr}
{ m{40em} } 
% {l}
\hline
Title: 'Pack Lego': Perth family caught in hard border crossfire at Christmas\\
\hline
Description: Perth mother Clare has found herself mostly confined to a small Sydney city apartment with her six-year-old and two-year-old as the northern beaches outbreak takes its toll.\\
\hline
Full text: Perth mother Clare* has found herself mostly confined to a small Sydney city apartment with her two young daughters as the northern beaches outbreak takes its toll. Clare flew with Cora, 6, and Georgia, 2, to Sydney on December 11 so they could reunite as a family over Christmas with husband Charles, who has been working remotely for the past year. Perth mother Clare and daughter Cora are in Sydney during the Christmas hard border closure. "It's pretty tough because we're in a really small apartment which is not set up for small kids," Clare said. 
"We worked on the theory that it'd be fine because we'd be out and about during the day doing tourist things in Sydney. But obviously we've tried to stay home as much as we can, apart from a grocery shop." Advertisement It's also been raining "a lot", which has put a damper on escaping to the big parks near their suburb of Haymarket that sits adjacent to tourism hotspot Darling Harbour.\\
\hline\\
\caption{Sample text used to find the 15 most significant words using the TF - IDF vectorizer.}
\label{fig:Table 3.6}
\end{tblr}
\end{table}

%% file: table_4.tex
\begin{table}
\centering
\resizebox{\linewidth}{!}{%
\begin{tblr}{
cell{1}{2} = {c = 2}{halign = c},
cell{1}{4} = {c = 2}{halign = c},
}
\hline
Model & 
OvR and Multinomial Naive Bayes & & 
K Neighbours Classifier &\\
\hline
Data Collection & 
Web scraping & Boilerplate removal & 
Web scraping & Boilerplate removal\\
\hline
Accuracy & 0.007 & 0.0063 & 0.4344 & 0.5165\\
Macro Recall & 0.0157 & 0.9918 & 0.8761 & 0.8664\\
Micro Recall & 0.441 & 0.8737 & 0.8629 & 0.8561\\
Weighted Recall & 0.441 & 0.8737 & 0.8629 & 0.8561\\
Macro F1 & 0.0136 & 0.1371 & 0.5071 & 0.5771\\
Micro F1 & 0.4569 & 0.6288 & 0.7759 & 0.7992\\
Weighted F1 & 0.4332 & 0.851 & 0.8229 & 0.8332\\
Macro Precision & 0.0653 & 0.1342 & 0.4876 & 0.5947\\
Micro Precision & 0.474 & 0.4911 & 0.7048 & 0.7494\\
Weighted Precision & 0.4547 & 0.8455 & 0.807 & 0.8387\\
\hline\\
\end{tblr}
}
\caption{Performance metrics for the One vs Rest Classifier and the K Neighbors Classifier.}
\label{fig:Table 4.1}
\end{table}